\documentclass[conference]{IEEEtran}
\usepackage{stmaryrd}
\usepackage{amsfonts}

\usepackage{graphicx}
\usepackage{tabularx}
\usepackage{setspace}

\usepackage{url}
\usepackage{amsmath}
\usepackage{amssymb}
\usepackage{array}
\usepackage{longtable}

\usepackage{float}
\usepackage{epsfig}
\usepackage{setspace}
\usepackage{multirow}
\usepackage{array}
\usepackage{subfigure}
\usepackage{wasysym}

\def\Hline{\noalign{\hrule height 4\arrayrulewidth}}
\newcolumntype{V}{>{$\vcenter\bgroup\hbox\bgroup}c<{\egroup\egroup$}}

\graphicspath{{../2013_AAAI1/imgs/},{./imgs/}}

\IEEEoverridecommandlockouts    

\begin{document}

\title{\LARGE\bf Illumination-Invariant Face Recognition from a Single Image across Extreme Pose using a Dual Dimension AAM Ensemble in the Thermal Infrared Spectrum}

\author{Reza Shoja Ghiass$^{\dag a}$, Ognjen Arandjelovi\'c$^{\ddag b}$, Hakim Bendada$^\dag$, and Xavier Maldague$^\dag$\\$^\dag$~Universit\'{e} Laval, Quebec, Canada ~~~~~~~~~~ $^\ddag$~Deakin University, Geelong, Australia\\\small$^a$~\texttt{reza.shoja@gmail.com} ~~~~~~~~~~~~~~~~~~~~~~~~~~~ $^b$~\texttt{ognjen.arandjelovic@gmail.com}}


\maketitle

\begin{abstract}
Over the course of the last decade, infrared (IR) and particularly thermal IR imaging based face recognition has emerged as a promising complement to conventional, visible spectrum based approaches which continue to struggle when applied in practice. While inherently insensitive to visible spectrum illumination changes, IR data introduces specific challenges of its own, most notably sensitivity to factors which affect facial heat emission patterns, e.g.\ emotional state, ambient temperature, and alcohol intake. In addition, facial expression and pose changes are more difficult to correct in IR images because they are less rich in high frequency detail which is an important cue for fitting any deformable model. In this paper we describe a novel method which addresses these major challenges. Specifically, when comparing two thermal IR images of faces, we mutually normalize their poses and facial expressions by using an active appearance model (AAM) to generate synthetic images of the two faces with a neutral facial expression and in the same view (the average of the two input views). This is achieved by piecewise affine warping which follows AAM fitting. A major contribution of our work is the use of an AAM ensemble in which each AAM is specialized to a particular range of poses \emph{and} a particular region of the thermal IR face space. Combined with the contributions from our previous work which addressed the problem of reliable AAM fitting in the thermal IR spectrum, and the development of a person-specific representation robust to transient changes in the pattern of facial temperature emissions, the proposed ensemble framework accurately matches faces across the full range of yaw from frontal to profile, even in the presence of scale variation (e.g.\ due to the varying distance of a subject from the camera). The effectiveness of the proposed approach is demonstrated on the largest public database of thermal IR images of faces and a newly acquired data set of thermal IR motion videos. Our approach achieved perfect recognition performance on both data sets, significantly outperforming the current state of the art methods even when they are trained with multiple images spanning a range of head views.
\end{abstract}


\section{Introduction}\label{s:intro}
\PARstart{N}{otwithstanding} substantial and continued research efforts, the practical success of face recognition using conventional imaging equipment has been limited. In an effort to overcome some of the key challenges which remain, such as appearance changes due to varying illumination and disguises (including facial wear and hair), the use of imaging outside the visible part of the electromagnetic spectrum (wavelength approximately in the range $390-750$~nm) has been explored. Most notably, the use of infrared data (wavelength approximately in the range $0.74-300$~$\mu$m) has been increasingly popular, in no small part due to the ever reducing cost of infrared cameras. For a recent comprehensive review please see \cite{GhiaAranBendMald2013b}.

In the literature, it has been customary to divide the IR spectrum into four sub-bands: near IR (NIR; wavelength $0.75-1.4\mu$m), short wave IR (SWIR; wavelength $1.4-3\mu$m), medium wave IR (MWIR; wavelength $3-8\mu$m), and long wave IR (LWIR; wavelength $8-15\mu$m). This division of the IR spectrum is also observed in the manufacturing of IR cameras, which are often made with sensors that respond to electromagnetic radiation constrained to a particular sub-band. It should be emphasized that the division of the IR spectrum is not arbitrary. Rather, different sub-bands correspond to continuous frequency chunks of the solar spectrum which are divided by absorption lines of different atmospheric gasses \cite{Mald2001}. In the context of face recognition, one of the largest differences between different IR sub-bands emerges as a consequence of the human body's heat emission spectrum. Specifically, most of the heat energy is emitted in LWIR sub-band, which is why it is often referred to as the thermal sub-band (this term is often extended to include the MWIR sub-band). Significant heat is also emitted in the MWIR sub-band. Both of these sub-bands can be used to \emph{passively} sense facial thermal emissions without an external source of light. This is one of the reasons why LWIR and MWIR sub-bands have received the most attention in the face recognition literature. In contrast to them, facial heat emission in the SWIR and NIR sub-bands is very small and systems operating on data acquired in these sub-bands require appropriate illuminators i.e.\ recognition is \emph{active} in nature. In recent years, the use of NIR also started to receive increasing attention from the face recognition community, while the utility of the SWIR sub-band has yet to be studied in depth.

\paragraph{Advantages of IR based face recognition}
The foremost advantage of IR data in comparison with conventional visible spectrum images in the context of face recognition lies in its invariance to visible spectrum illumination. This is inherent in the very nature of IR imaging and, considering the challenge that variable illumination conditions present to face recognition systems, a major advantage. What is more, IR energy is also less affected by scattering and absorption by smoke or dust than reflected visible light \cite{ChanKoscAbidKong+2008a,NicoSchm2011}. Unlike visible spectrum imaging, IR imaging can be used to extract not only exterior but also useful subcutaneous anatomical information, such as the vascular network\footnote{It is important to emphasize that none of the existing publications on face recognition using `vascular network' based representations provide any evidence that the extracted structures are indeed blood vessels. Thus the reader should understand that we use this term for the sake of consistency with previous work, and that we do \emph{not} claim that what we extract in this paper is an actual vascular network. Rather we prefer to think of our representation as a \emph{function} of the underlying vasculature.} of a face \cite{BuddPavlTsiaBaza2007} or its blood perfusion patterns \cite{WuWeiFangLi+2007}. Finally, thermal vision can be used to detect facial disguises \cite{PavlSymo2000} as well.

\paragraph{Challenges in IR based face recognition}
The use of IR images for AFR is not void of its problems and challenges. For example, MWIR and LWIR images are sensitive to the environmental temperature, as well as the emotional, physical and health condition of the subject. They are also affected by alcohol intake. Another potential problem is that eyeglasses are opaque to the greater part of the IR spectrum (LWIR, MWIR and SWIR) \cite{TasmJaeg2009}. This means that a large portion of the face wearing eyeglasses may be occluded, causing the loss of important discriminative information. The complementary nature of visible spectrum images in this regard has inspired various multi-modal fusion methods \cite{HeoKongAbidAbid2004,AranHammCipo2006,AranHammCipo2010}. Another consideration of interest pertains to the impact of sunlight if recognition is performed outdoors and during daytime.  Although invariant to the changes in the illumination by visible light itself (by definition), the IR ``appearance'' in the NIR and SWIR sub-bands \emph{is} affected by sunlight which has significant spectral components at the corresponding wavelengths. This is one of the key reasons why NIR and SWIR based systems which perform well indoors struggle when applied outdoors \cite{LiChuLiaoZhan2007}.

\paragraph{Previous work}
The earliest attempts at examining the potential of infrared imaging for face recognition date back to the work done by Prokoski \textit{et al.}\ \cite{ProkRiedCoff1992}. Most of the automatic methods which followed closely mirrored the methods developed for visible spectrum based recognition. Generally, these used holistic face appearance in a simple statistical manner, with little attempt to achieve any generalization, relying instead on the availability of training data with sufficient variability of possible appearance for each subject \cite{Cutl1996,SocoWolfNeuhEvel2001,SeliSoco2004}. More sophisticated holistic approaches recently investigated include statistical models based on Gaussian mixtures \cite{ElguBoug2011} and compressive sensing \cite{LinWenrLiZhij2011}. Numerous feature based approaches have also been described. The use of locally binary patterns was proposed by Li \textit{et al.}\ \cite{LiChuLiao+2007} and Goswami \textit{et al.}\ \cite{GoswChanWindKitt2011}, scale invariant transform (SIFT) features by Maeng \textit{et al.}\ \cite{MaenChoiParkLee+2011}, wavelets by Srivastava and Liu \cite{SrivLiu2003}, and Nicolo and Schmid \cite{NicoSchm2011}, and curvelets by Xie \textit{et al.}\ \cite{XieWuLiuFang2009a}. The method of Wu \textit{et al.}\ \cite{WuSongJianXie+2005} is one of the few in the literature which attempts to extract useful subcutaneous information from IR appearance. Using a blood perfusion model Wu \textit{et al.}\ infer the blood perfusion pattern corresponding to an IR image of a face. Buddharaju \textit{et al.}\ \cite{BuddPavlTsia2005} instead extract the vascular network of a face. They represent vascular networks as binary images (each pixel either is or is not a part of the network) and match them using a method adopted from fingerprint recognition: using salient loci of the networks (such as bifurcation points). While successful in the recognition of fingerprints which are planar, this approach is not readily adapted to deal with pose changes expected in many practical applications of face recognition. In addition, as we will explain in further detail in Sec.~\ref{sss:vesselness}, the binary nature of their method for vascular network extraction makes it sensitive to face scale and image resolution.

\section{Dual Dimension AAM Ensemble}

\subsection{Proposed approach overview}
Considering the inherent insensitivity of IR images to visible illumination changes, in this work we focus on achieving robustness with respect to three key remaining nuisance factors: (i) transient physiological changes that affect facial thermal emissions, (ii) pose, and (iii) facial expression. The robustness with respect to transient physiological changes is achieved by the use of an invariant, person-specific representation based on the distribution of superficial blood vessels in the face adopted from our previous work \cite{GhiaAranBendMald2013}, while pose and facial expression changes are normalized using an appearance model (AAM) \cite{CootEdwaTayl1998} ensemble framework. This framework comprises two components, each of which is essential for ensuring that an AAM is fitted correctly across the full range of poses, from frontal to full profile. Firstly, we do not perform fitting on raw thermal IR images, which lack characteristic detail needed to constrain and guide the fitting process. Instead, we pre-process images so as to emphasize important discriminative information. This improves the performance of AAM fitting but still does not solve the problem entirely: even for head orientations close to frontal the model often fails to converge to the vicinity of the globally optimal parameters, with a further increase in the incidence of failures as the pose becomes more extreme. As a solution we employ not a single AAM but rather an ensemble of AAMs. Each AAM is trained on an automatically determined cluster of thermal IR images in the training database, where clustering is performed both across individuals and poses. In other words, each AAM corresponds to a particular, constrained range of poses and includes the images of people who look similar across these views. Each of the components of our system is described in detail next, starting with a review of the inverse compositional AAM, its modification proposed here and used as the baseline fitting procedure for a single AAM, followed by its contextualization within the proposed ensemble framework.

\subsection{Pose and facial expression normalization using the AAM}
Much like in the visible spectrum, the appearance of a face in the thermal IR spectrum is greatly affected by the person's head pose relative to the camera~\cite{FrieYesh2003}. Therefore it is essential that recognition is performed either using features which are invariant to pose changes or that pose is synthetically normalized. In the present paper we adopt the latter approach. Specifically, given two images of faces that we wish to compare, we first fit an AAM to each, and then synthetically generate two images of the faces in the same pose in which they can be readily compared. Although widely used for visible spectrum based face recognition \cite{FengShenZhouZhang+2011,SaueCootTayl2011}, to the best of the knowledge of these authors, ours is the first published work which has attempted to apply it on IR data (specifically thermal IR data) \cite{GhiaAranBendMald2013}. Unlike previous work (including ours \cite{GhiaAranBendMald2013}), we do not necessarily synthesize images of faces in the frontal pose. Rather, to minimize the amount of distortion produced by the approximative nature of the AAM which comprises piece-wise linear surface patches, we warp two faces to the canonical frame corresponding to roughly the pose between the initial input poses. This is facilitated by our ensemble framework, covered in detail in Sec.~\ref{ss:ensemble}. For now, we start with a description of the proposed baseline AAM fitting adopted from \cite{GhiaAranBendMald2013} where it was recently first described.

\subsubsection{Inverse compositional AAM fitting}
The type of an active appearance model we are interested in here separately models the face shape, as a piecewise triangular mesh, and face appearance, covered by the mesh \cite{CootEdwaTayl1998}. The model is trained using a data corpus of faces which has salient points manually annotated and which become the vertices of the corresponding triangular mesh. These training meshes are used to learn the scope of variation of shape (i.e.\ locations of vertices) and appearance of individual mesh faces, both using principal component analysis i.e.\ by retaining the first principal components as the basis of the learnt generative model. The model is applied on a novel face by finding a set of parameters (shape and appearance principal component weights) such that the difference between the corresponding piecewise affine warped image and the model predicted appearance is minimized. Formally, the fitting error can be written as:
{\begin{align}
  e_{aam} = \sum_{\text{All pixels } \mathbf{x}} \left[ A_0(\mathbf{x})+\sum_{i=1}^m \alpha_i A_i(\mathbf{x}) - I_e(\mathbf{W}(\mathbf{x};\mathbf{p})) \right]^2,
  \label{e:aam}
\end{align}}
where $\mathbf{x}$ are pixel loci, $A_i$ ($i=0\ldots m$) the retained appearance principal components and $\mathbf{W}(\mathbf{x};\mathbf{p})$ the location of the pixel warped using the shape parameters $\mathbf{p}$. For additional detail, the reader is referred to the original publication \cite{CootEdwaTayl1998}.

The most straightforward approach to minimizing the error described in Eq.~\ref{e:aam} is by using gradient descent. However, this is slow. A popular alternative proposed by Cootes and Taylor \cite{CootEdwaTayl1998} uses an approximation of a linear relationship between the fitting error term in Eq.~\ref{e:aam}, and the updates to the shape and appearance parameters, respectively $\Delta \mathbf{p}$ and $\Delta \alpha_i$ (more complex variations on the same theme include \cite{SclaIsid1998,CootEdwaTayl2001}). What is more, the linear relationship is assumed not to depend on the model parameters, facilitating a simple learning of the relationship from the training data corpus. While much faster than gradient descent, this fitting procedure has been shown to produce inferior fitting results in comparison to the approach we adopt here, the inverse compositional AAM (ICAAM) \cite{MattBake2004}. Our experiments show that the advantages of ICAAM are even greater when fitting is done on thermal IR images, as opposed to visual ones (used by all of the aforementioned authors).

There are two keys differences between the conventional AAM fitting and ICAAM. Firstly, instead of estimating a simple update of the parameters $\Delta \mathbf{p}$, the compositional AAM algorithm estimates the update to the warp itself, i.e.\ $\mathbf{W}(\mathbf{x},\mathbf{p})$. This particular idea was first advanced by Lucas and Kanade \cite{LucaKana1981}. Secondly, in the inverse compositional algorithm, the direction of piecewise linear warping is inverted. Instead of warping the input image to fit the reference mesh, the warping is performed in the opposite direction. In other words, the error minimized becomes:
{\begin{align}
  e_{icaam} = \sum_{\text{All pixels } \mathbf{x}} \left[ I_e(\mathbf{W}(\mathbf{x};\mathbf{p})) - A_0(\mathbf{W}(\mathbf{x};\mathbf{p})) \right]^2.
  \label{e:icaam}
\end{align}}
While it can be shown that the minimization of Eq.~\ref{e:icaam} is equivalent to that of Eq.~\ref{e:aam}, this inverse formulation allows much of the intensive computation to be pre-computed, thus resulting in faster convergence in addition to the increased fitting accuracy provided by the compositional AAM.

\subsection{Preprocessing}\label{ss:preprocessing}

\subsubsection{Face segmentation}
One of the key problems encountered in practical application of AAM is the initialization. If the initial geometric configuration of the model is too far from the correct solution, the fitting may converge to a locally optimal but incorrect set of values (loci of salient points). This is a particularly serious concern for thermal IR images since thermal appearance of faces is less rich in detail which guides and constrains the AAM. For this reason, in our method face segmentation is performed first. This accomplishes two goals. Firstly, the removal of any confounding background information helps the convergence in the fitting of the model. Secondly, the AAM can be initialized well, by considering the shape and scale of the segmented foreground region.

\begin{figure*}[htb]
  \centering
  \subfigure[]{\includegraphics[width=0.2\textwidth,trim=10mm 12mm 15mm 8mm,clip]{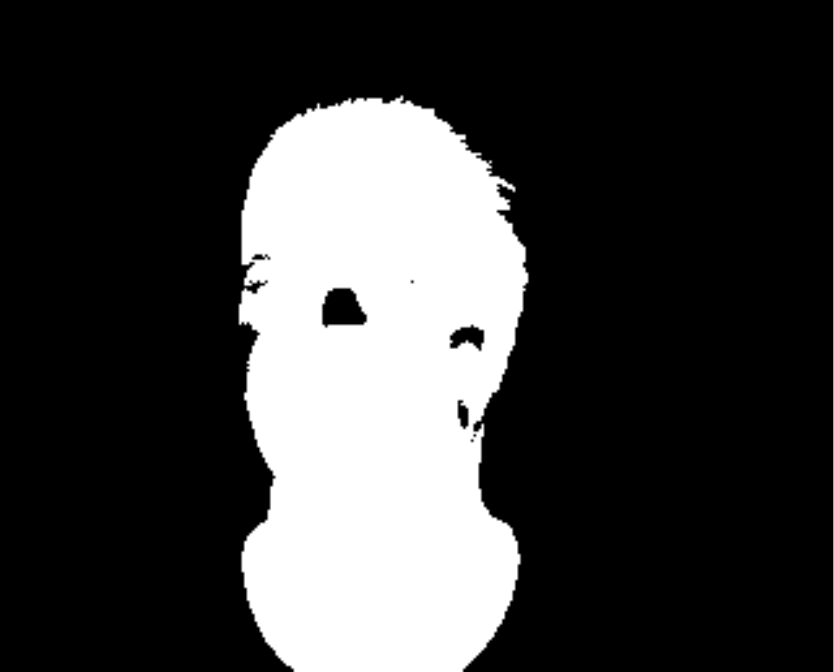}}~~~~~~
  \subfigure[]{\includegraphics[width=0.2\textwidth,trim=10mm 12mm 15mm 8mm,clip]{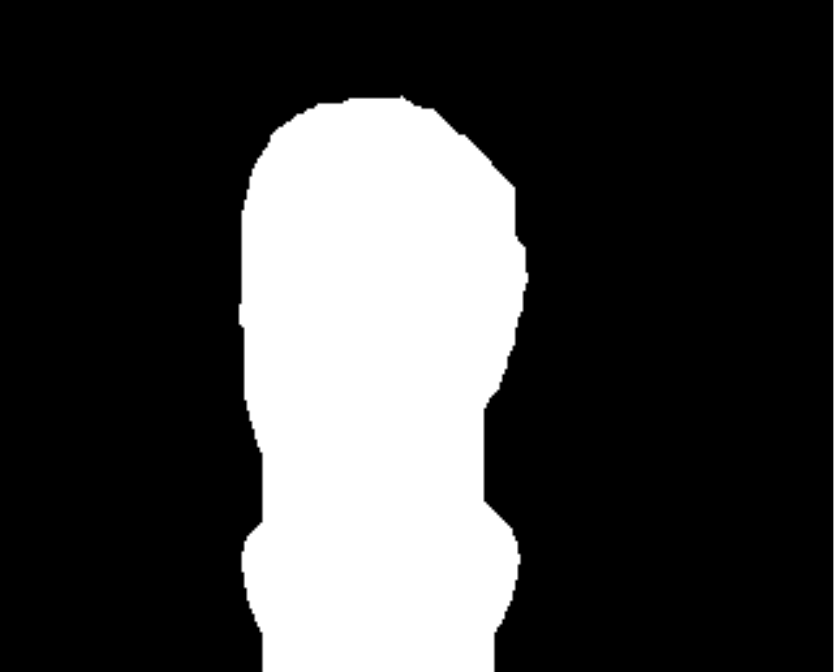}}~~~~~~
  \subfigure[]{\includegraphics[width=0.2\textwidth,trim=10mm 12mm 15mm 8mm,clip]{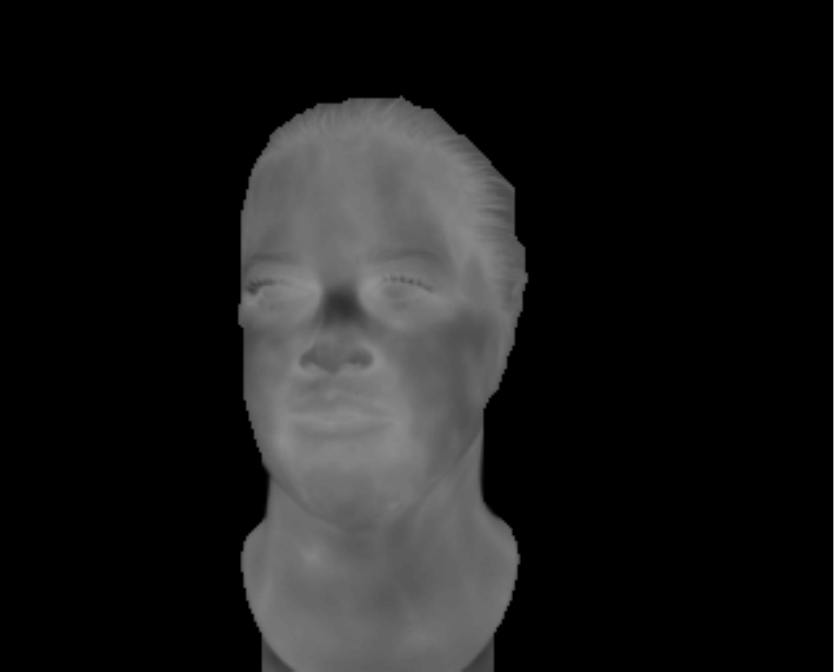}}
  \caption{ The first step in the proposed algorithm is to segment out the face. This removed image areas unrelated to the subject's identity and aids in the convergence of the AAM. From (a) the original image (b) provisional segmentation mask is created using temperature thresholding, after which (c) morphological operators are used to increase the segmentation accuracy to outliers e.g.\ such as which occur at the interface of facial and non-facial regions, producing (d) the   final result with the background correctly suppressed.}
  \label{f:segm}
\end{figure*}

Unlike in the visible spectrum, in which background clutter is often significant and in which face segmentation can be a difficult task, face segmentation in thermal IR images is in most cases far simpler. Indeed, in the present paper we accomplish the bulk of work using simple thresholding. We create a provisional segmentation map by declaring all pixels with values within a range between two thresholds, $T_{low}$ and $T_{up}$, as belonging to the face region (i.e.\ foreground) and all others as background. An example is shown in Fig.~\ref{f:segm}(a). The provisional map is further refined by performing morphological opening and closing operations, using a circular structuring element (we used a circle whose approximate area is 6\% of the area of the segmented face ellipse). This accomplishes the removal of spurious artefacts which may occur at the inferface between the subject's skin and clothing for example, and correctly classifies facial skin areas which are unusually cold or hot (e.g.\ the exposure to cold surroundings can transiently create cold skin patches in the regions of poor perifocal vascularity). An example of the final segmentations mask is shown in Fig.~\ref{f:segm}(c) and the segmented image output in Fig.~\ref{f:segm}(d).

\subsubsection{Detail enhancement}
\begin{figure}[htb]
  \centering
  \subfigure[Diffused]{\includegraphics[width=0.2\textwidth,trim=10mm 12mm 15mm 8mm,clip]{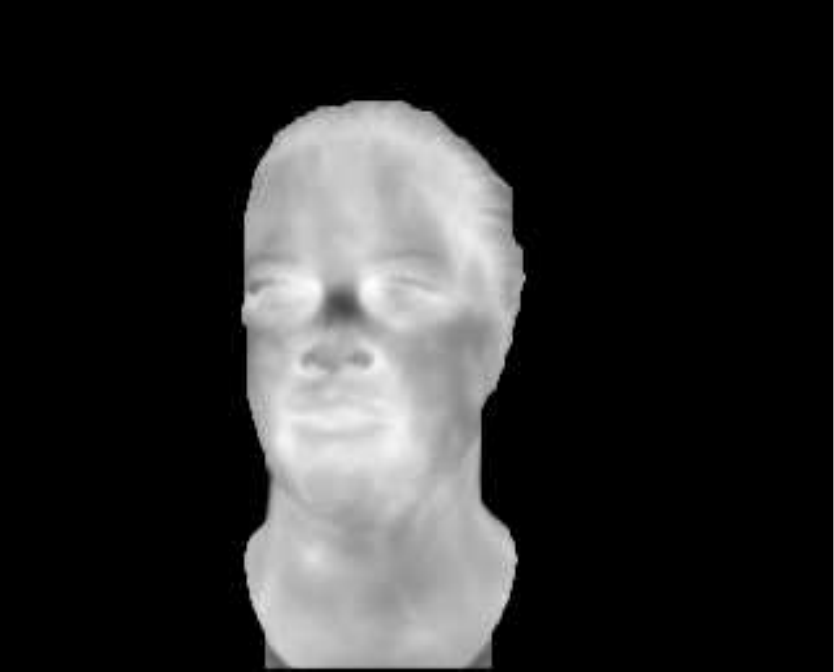}}~~~~~~
  \subfigure[Enhanced]{\includegraphics[width=0.2\textwidth,trim=10mm 12mm 15mm 8mm,clip]{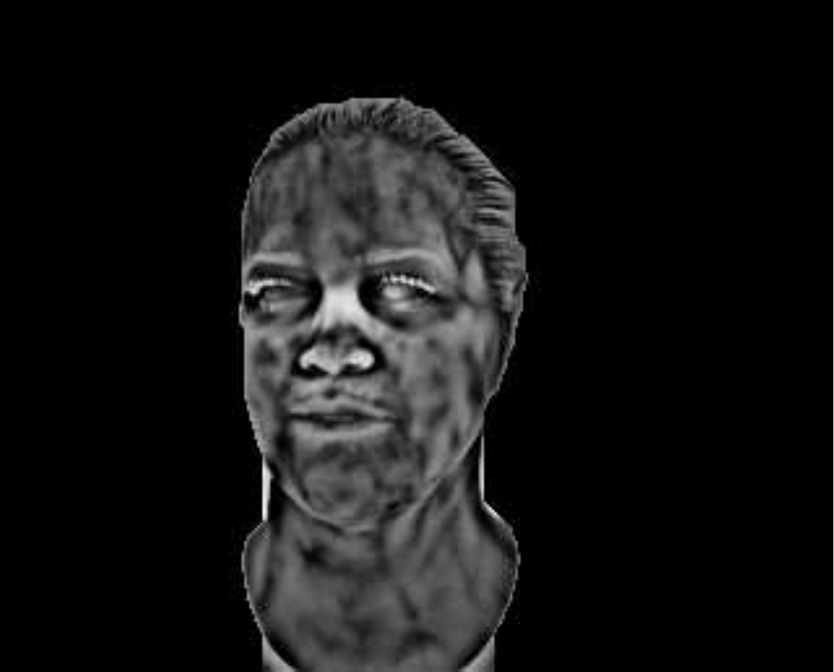}}
  \vspace{3pt}
  \caption{ The AAM is notoriously sensitive to initialization. This potential problem is even greater when the model is used on thermal images, which lack characteristic, high frequency content. We increase the accuracy of AAM fitting first by (a) creating an anisotropically smoothed thermal image, which is then (b) subtracted from the original image to produce an image with enhanced detail. }
  \label{f:enh}
\end{figure}

As mentioned earlier, thermal IR images of faces are much less rich in fine detail than visible spectrum images \cite{ChenFlynBowy2003}, which be readily observed in the example in Fig.~\ref{f:enh}(a). This makes the problem of AAM fitting all the more challenging. For this reason, we do not train or fit the AAM on segmented thermal images themselves, but on processed and detail enhanced images. In our experiments we found that the additional detail the proposed filtering brings out greatly aids in guiding the AAM towards the correct solution.

The method for detail enhancement we adopt from our previous work \cite{GhiaAranBendMald2013} is a form of non-linear high pass filtering. Firstly, we anisotropically diffuse the input image using diffusion of the form:
{\begin{align}
  \frac{\partial I}{\partial t} = \nabla.\left( c(\| \nabla I\|)~\nabla I\right) = \nabla c.\nabla I + c(\| \nabla I\|)~\Delta I,
\end{align}}
with the diffusion parameter $c$ constant over time (i.e.\ filtering iterations) but spatially varying and dependent on the magnitude of the image gradient:
{\begin{align}
  c(\| \nabla I\|) = \exp \left\{ -\frac{\|\nabla I\|} {k^2} \right\}.
\end{align}}
We used $k=20$. The detail enhanced image is then computed by subtracting the diffused image from the original: $I_e = I - I_d$. This is illustrated on a typical example in Fig.~\ref{f:enh}(a,b).

\subsection{AAM ensemble framework}\label{ss:ensemble}
The baseline AAM fitting approach described in Sec.~\ref{ss:preprocessing} greatly improves the accuracy of model parameter estimates after convergence. On nearly frontal faces (yaw deviation of up to $\pm30^\circ$ from frontal), when raw thermal IR images are used on our databases (see Sec.~\ref{ss:data}) we found that the AAM converges to an acceptable solution in fewer than 40\% of the cases. In contrast, our approach results is successful for all nearly frontal images in which the person does not have substantial facial hair (beard) and is not wearing eyeglasses. However, as the head orientation of the person in an input image becomes further from frontal, even this method exhibits a deterioration of performance which becomes unacceptable for poses close to full profile. To solve this problem we propose what we term a Dual Dimension AAM Ensemble (DDAE).

\begin{figure}[htb]
  \centering
  \includegraphics[width=0.48\textwidth]{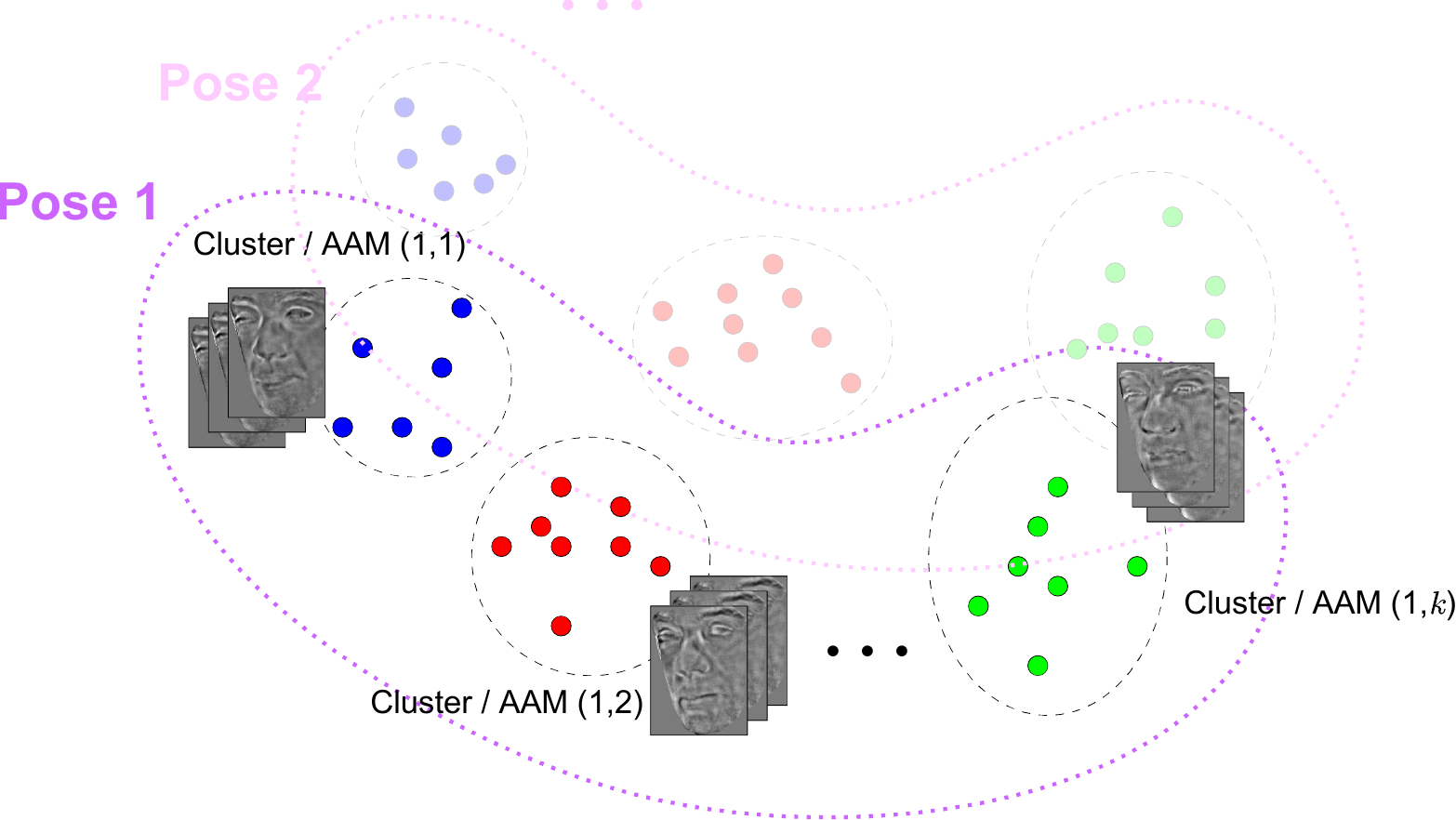}
  \vspace{3pt}
  \caption{ A conceptual illustration of the proposed Dual Dimension AAM Ensemble. Training data used for different AAMs is first separated by restricting the range of head poses; then, training data within each head pose range is clustered, resulting in the specialization of each AAM to a region of the thermal IR face space corresponding to specific poses and facial appearances. }
  \label{f:ensembleoverview}
\end{figure}

The key idea behind the proposed method is to employ not a single, generic AAM but rather an ensemble of AAMs, each `specializing' in a particular range of modelled variation. We restrict both pose and inter-personal variation covered by each AAM (thus `Dual Dimension'). First, we divide the training data set into subsets, each of which corresponds to a specific pose range. In this paper we used three yaw ranges (i.e.\ three training data subsets): (i) $0-45^\circ$, (ii) $22.5-67.5^\circ$, and (iii) $45-90^\circ$. Next, we performed data clustering on each subset. The goal of this step is to mine the data subsets for individuals with similar thermal IR appearance in a particular range of views, thus allowing the specialization of an AAM trained only on this particular cluster. Consequently, if $n_{\text{pose}}$ is the number of distinct pose ranges that covers the entirety of pose variation from frontal to profile, and $n_{\text{AAM}}^{(i)}$ the number of people clusters discovered within that pose range, the total number of AAMs trained is:
\begin{align}
  n_{\text{AAM}} = \sum_{i=1}^{n_{\text{pose}}} n_{\text{AAM}}^{(i)}.
\end{align}
This is illustrated conceptually in Fig.~\ref{f:ensembleoverview}. For the sake of simplicity, in the present paper we fixed the number of clusters per pose so that:
\begin{align}
  \forall i.~n_{\text{AAM}}^{(i)} \equiv k.
\end{align}
We used $k=6$, resulting in $3\times 6=18$ AAMs in total. The use of more complex clustering methods which automatically discover the number of clusters in data is something that we wish to explore in our future work (see Sec.~\ref{s:conc}), particularly in the context of the method's scalability to very large data sets.

\paragraph{AAM selection} Since a novel input image contains a face in an arbitrary and unknown pose, a problem that emerges is that of selecting the appropriate AAM from our ensemble. Fortunately, the nature of the ensemble, that is, the specificity of each AAM leads to a simple and elegant solution. The fitting of an individual AAM can be seen as the maximization of the model's likelihood with respect to its free parameters (i.e.\ weights corresponding to its appearance and shape basis). This allows us to select the best AAM in the ensemble as the one with the greatest maximal likelihood, that is, the likelihood achieved after convergence. If the error of the AAM modelling the variations of the $i$-th pose range for the $j$-th people cluster is $e_{icaam}^{(i,j)}$, as per Eq.~\eqref{e:icaam}, then the AAM selected for the image $I$ is:
\begin{align}
  (i^*,j^*) = \arg \min_{(i,j)} e_{icaam}^{(i,j)}.
\end{align}

\subsection{Implementation and specific design considerations}
\begin{figure}[htb]
  \centering
  \subfigure[Typical AAM mesh design used in the visible spectrum]{\includegraphics[width=0.2\textwidth]{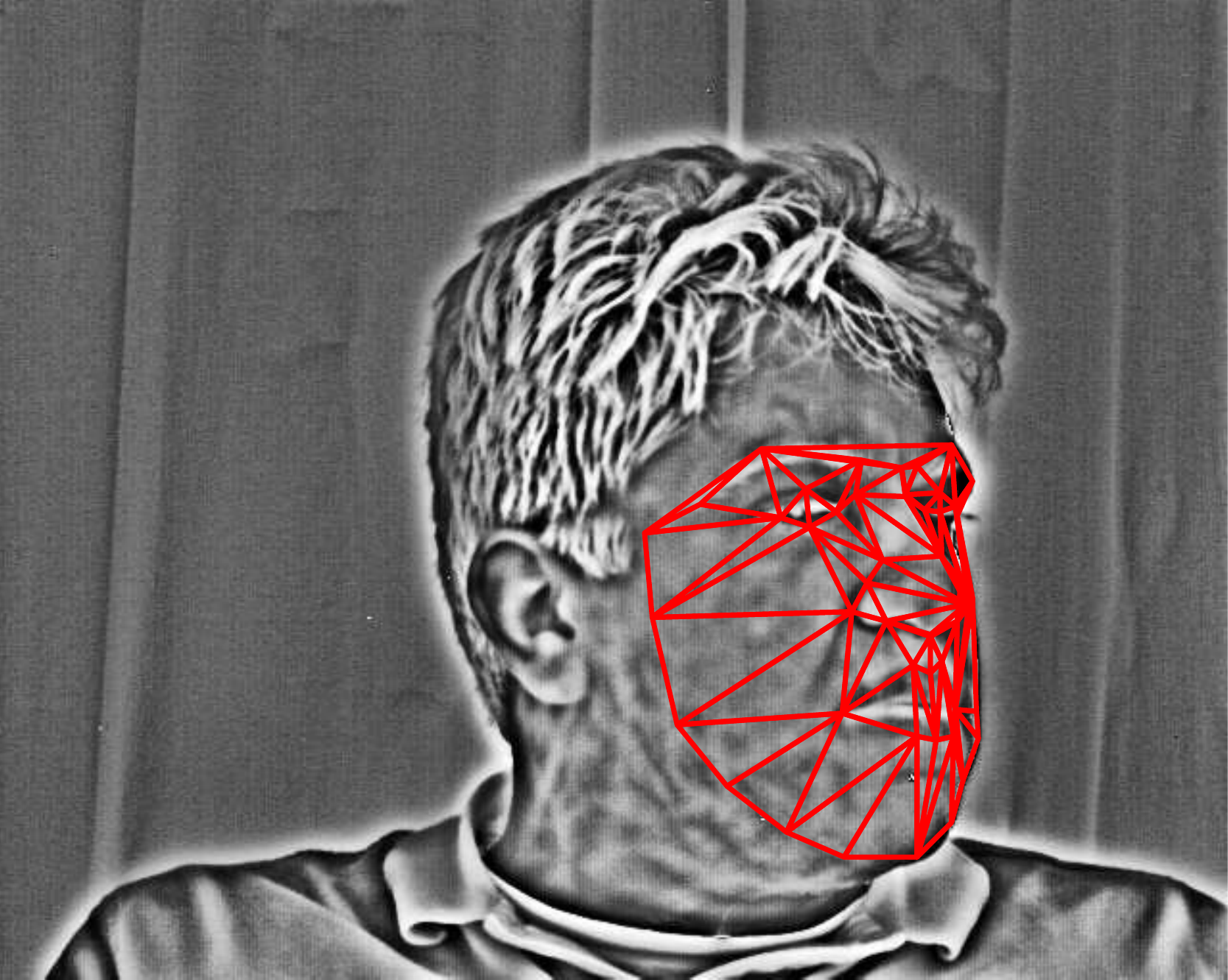}}~~~~~~
  \subfigure[AAM mesh used on IR images in the present paper]{\includegraphics[width=0.2\textwidth]{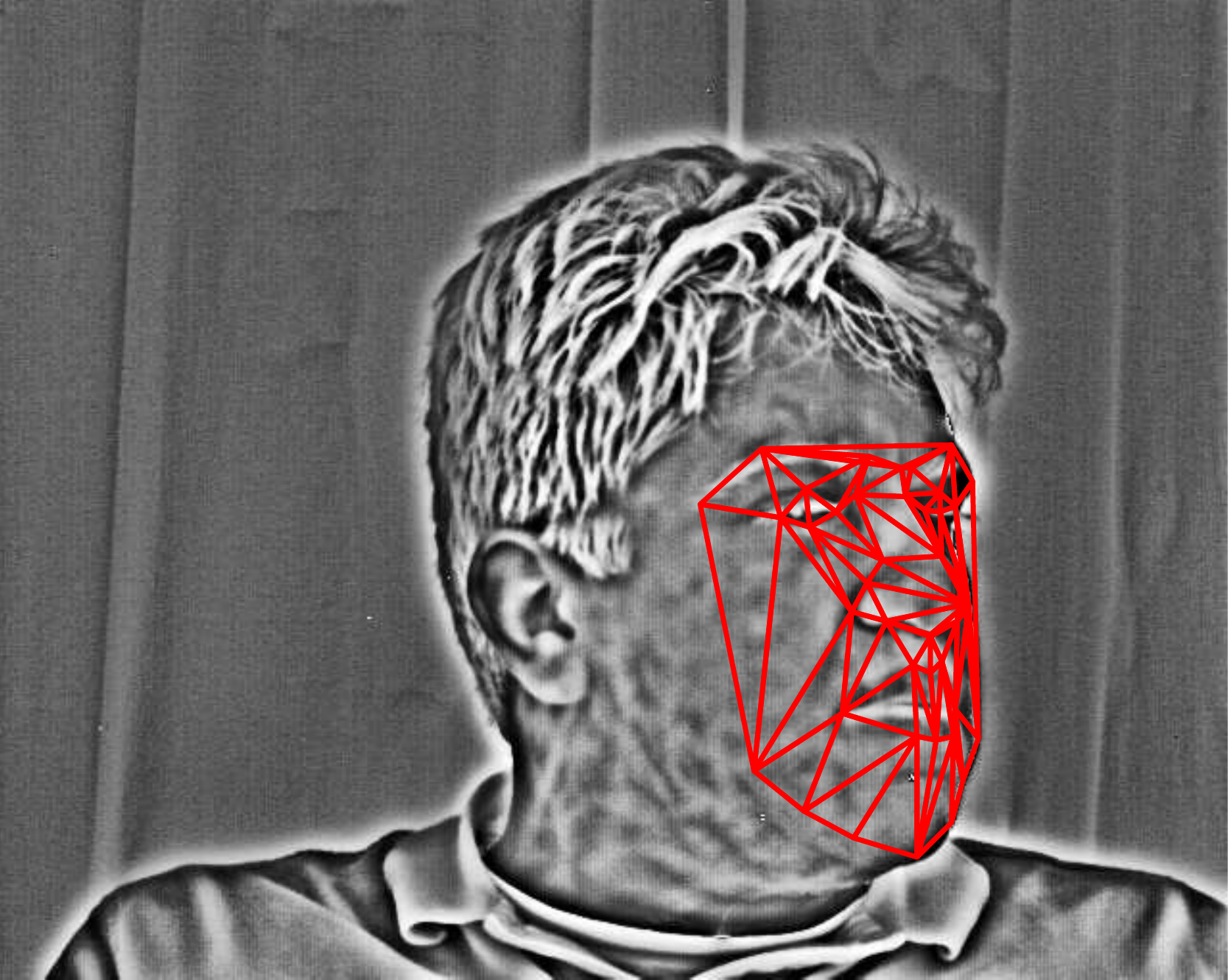}}
  \caption{ A comparison of (a) a typical AAM mesh design used for face recognition and tracking in the visible spectrum, and (b) the mesh design proposed in this paper. Notice that the placement of peripheral vertices in our design is closer to the frontal parts of the face, which lie on a closer to planar surface and which contain the greatest amount of discriminative information. This shift of the mesh boundary facilitates a more constrained mesh, in the sense that a greater number of mesh triangles cover nearly planar surfaces (which undergo simple appearance transformation with out of plane rotation) and a greater proportion of mesh vertices are placed on characteristic facial features. }
  \label{f:mesh}
\end{figure}

The last issue we would like to address with regard to the proposed AAM ensemble framework concerns the design of the model's triangular mesh. We found that the placement of mesh vertices as well as the triangulation over a fixed set of vertices can both have a dramatic effect on the fitting reliability. We are not aware of any in-depth treatment of this issue (which may be a contributing factor, or indeed a consequence, of the lack of AAM use in the IR based face recognition literature). Given its importance in the development of practical systems as well as in the reproducibility of the present work, we considered it worthwhile to highlight the differences of our mesh design compared to the common designs adopted when AAMs are used on images of faces acquired in the visible spectrum. A comparison is shown in Fig.~\ref{f:mesh}. Notice that the placement of peripheral vertices is closer to the frontal parts of the face. This allowed us to design a more constrained mesh, in the sense that a greater number of mesh triangles cover nearly planar surfaces (which undergo simple appearance transformation with out of plane rotation) and a greater proportion of mesh vertices are located on characteristic facial features.

\subsection{Discriminative representation}\label{ss:vesselEx}
Following the application of piecewise affine warping of the input image using ICAAM, all training and query faces are normalized for pose variations and the corresponding synthetically generated images contain frontal faces. Our goal in this stage of our algorithm is to extract from these images a person-specific representation which is firstly invariant to temperature changes and secondly robust to small imperfections of the preceding pose normalization.

\subsubsection{Differences from previous work}
As already argued in Sec.~\ref{s:intro}, the absolute temperature value of a particular point on a person's face can greatly vary as the conditions in which IR data is acquired are changed. The \emph{relative} temperature of different regions across the face is equally variable -- even simple physiological changes, such as an increase in sympathetic nervous system activity are effected non-uniformly. These observations strongly motivate the development of representations which are based on invariable anatomical features, which are unaffected by the aforementioned changeable distributions in heat emission.

One of the most prominent methods which has explored the use of invariant anatomical features was developed by Buddharaju \textit{et al.}\ \cite{BuddPavlTsiaBaza2007}. The key observation behind their method is that blood vessels are somewhat warmer than the surrounding tissues, allowing them to be identified in thermograms. These temperature differences are very small and virtually imperceptible to the naked eye, but inherently maintained regardless of the physiological state of the subject. An important property of vascular networks  which makes them particularly attractive for use in recognition is that the blood vessels are ``hardwired'' at birth and form a pattern which remains virtually unaffected by factors such as aging, except for predictable growth \cite{PersBusc2011}.

\begin{figure}[htb]
  \centering
  \small Vascular network of Buddharaju \textit{et al.}\\
  \subfigure[100\%]{\includegraphics[width=0.10\textwidth]{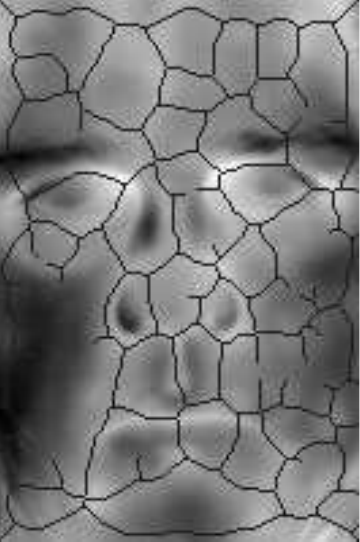}}~~
  \subfigure[ 90\%]{\includegraphics[width=0.10\textwidth]{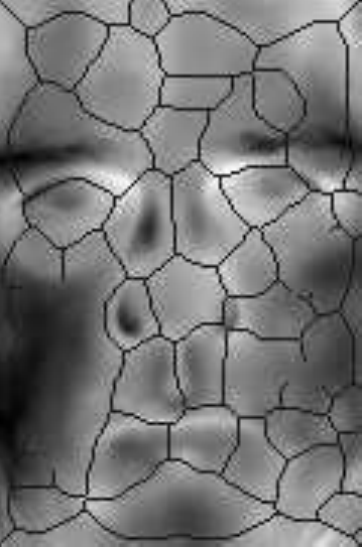}}~~
  \subfigure[ 80\%]{\includegraphics[width=0.10\textwidth]{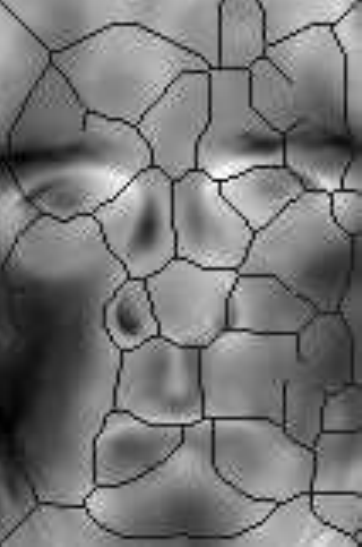}}~~
  \subfigure[ 70\%]{\includegraphics[width=0.10\textwidth]{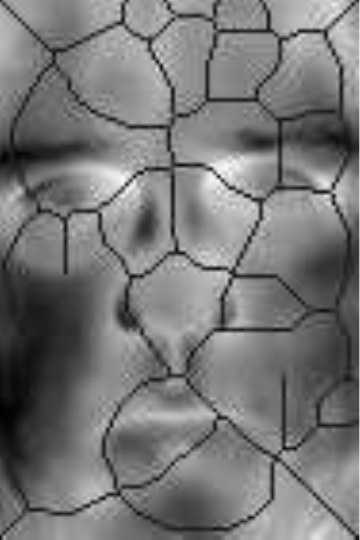}}
  \\
  \small Proposed vesselness response based representation\\
  \subfigure[100\%]{\includegraphics[width=0.10\textwidth]{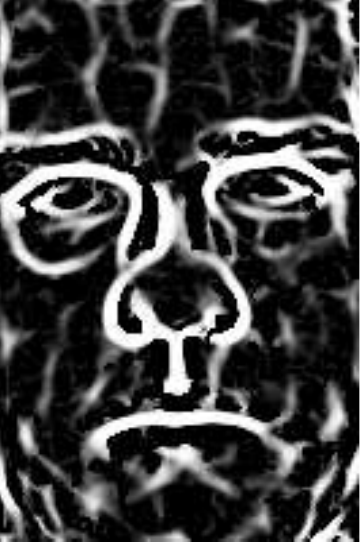}}~~
  \subfigure[ 90\%]{\includegraphics[width=0.10\textwidth]{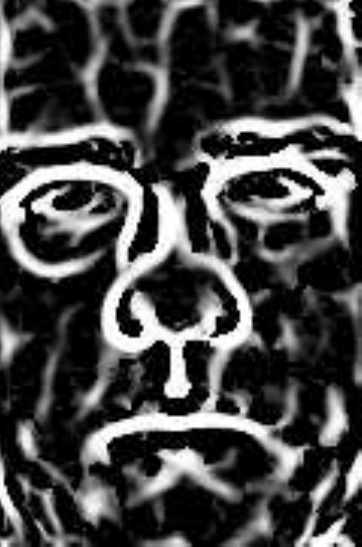}}~~
  \subfigure[ 80\%]{\includegraphics[width=0.10\textwidth]{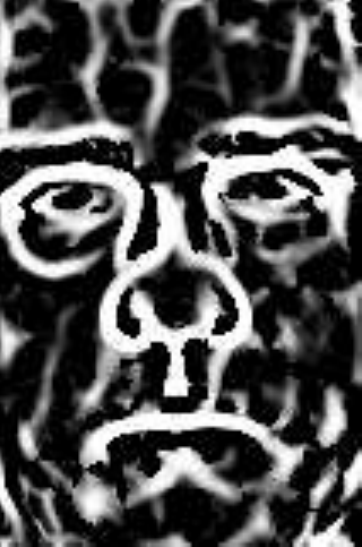}}~~
  \subfigure[ 70\%]{\includegraphics[width=0.10\textwidth]{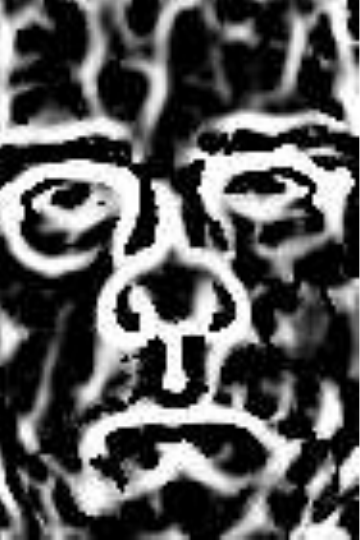}}
  \caption{ One of the major limitations of the vascular network based approach proposed by Buddharaju \textit{et al.} lies in its `crisp' binary nature: a particular pixel is deemed either a part of the vascular network or not. The consequence of this is that the extracted vascular network is highly sensitive to the scale of the input image (and thus to the distance of the user from the camera as well as the spatial resolution of the camera). (a-d) Even small changes in face scale can effect large topological changes on the result (note that the representation of interest is the vascular network, shown in black, which is only superimposed on the images it is extracted from for the benefit of the reader). (e-h) In contrast, the proposed vesselness response based representation encodes the certainty that a particular pixel locus is a reliable vessel pattern, and exhibits far greater resilience to scale changes.}
  \label{f:scale}
\end{figure}

In the present work we too adopt a vascular network based approach, but with several important differences in comparison with the previously proposed methods. The first of these is to be found in the manner vascular structures are extracted. Buddharaju \textit{et al.}\ adopt the use of a simple image processing filter based on the `top hat' function. We found that the output produced by this approach is very sensitive to the scale of the face (something not investigated by the original authors) and thus lacks robustness to the distance of the user from the camera. In contrast, the approach taken in the present paper is specifically aimed at extracting vessel-like structures, and it does so in a multi-scale fashion, integrating evidence across scales, see Fig.~\ref{f:scale}. The second major difference lies in the form in which the extracted network is represented. Buddharaju \textit{et al.}\ produce a binary image, in which each pixel is deemed either as belonging to the network or not, without any accounting for the uncertainty associated with this classification. This aspect of their approach makes it additionally sensitive to variable pose and temperature changes across the face (see \cite{ChenJuDingLiu2011} for related criticism). In contrast, in our baseline representation each pixel is (in principle) real-valued, its value quantifying the degree of confidence ($\in [0,1]$) that it is a part of the vascular structure.

\subsubsection{Vesselness}\label{sss:vesselness}
\begin{figure*}[thp]
  \centering
  \subfigure[Vesselness $\mathcal{V}(s)$ at the scale $s=3$ pixels]{\includegraphics[width=0.19\textwidth,trim=10mm 12mm 15mm 8mm,clip]{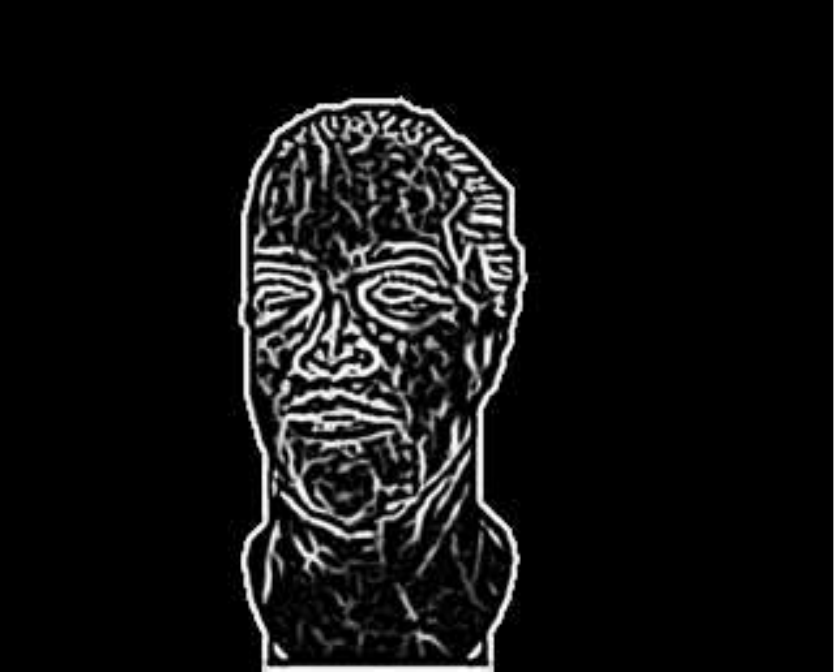}}~~~~
  \subfigure[Vesselness $\mathcal{V}(s)$ at the scale $s=4$ pixels]{\includegraphics[width=0.19\textwidth,trim=10mm 12mm 15mm 8mm,clip]{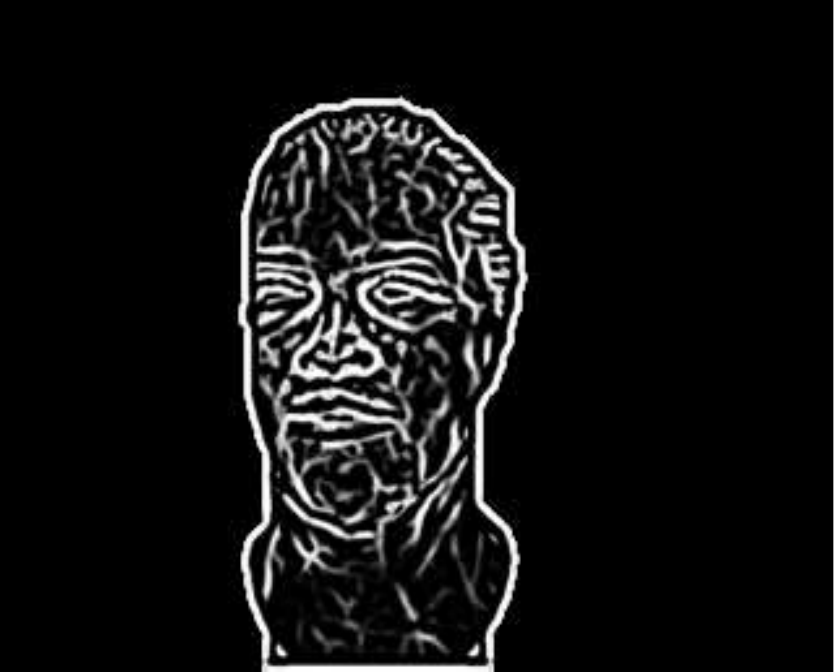}}~~~~
  \subfigure[Vesselness $\mathcal{V}(s)$ at the scale $s=5$ pixels]{\includegraphics[width=0.19\textwidth,trim=10mm 12mm 15mm 8mm,clip]{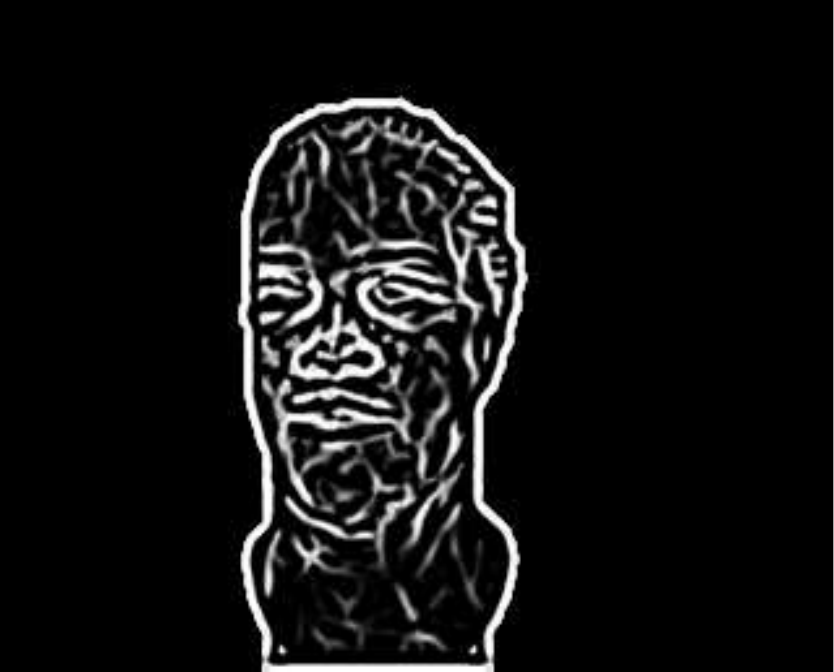}}~~~~
  \subfigure[Multi-scale vesselness $\mathcal{V}_0$ for $3\leq s\leq5$ pixels]{\includegraphics[width=0.19\textwidth,trim=10mm 12mm 15mm 8mm,clip]{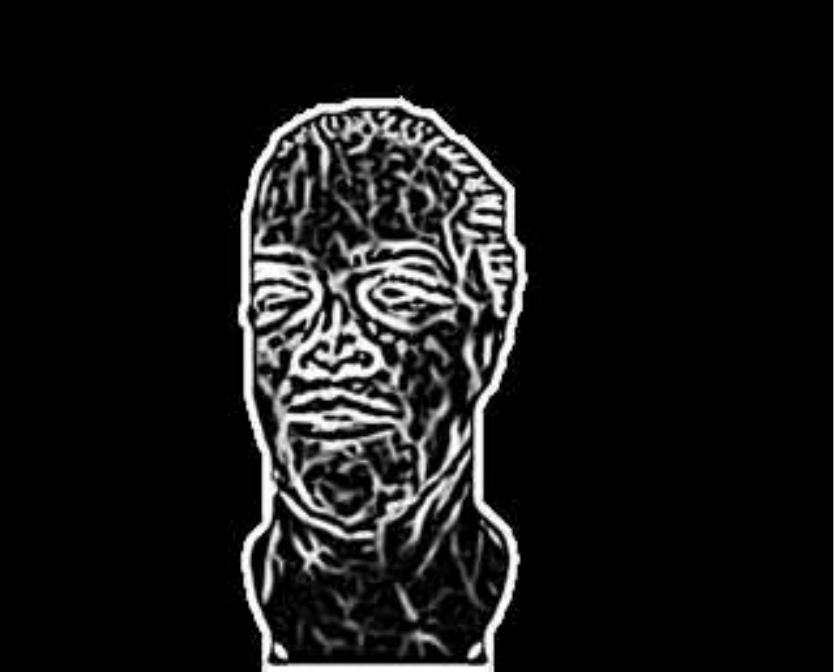}}
  \caption{ (a-c) The output of the vesselness filter at three different scales, and (d) the corresponding integrated multi-scale result. }
  \label{f:frangi}
\end{figure*}

We extract characteristic anatomical features of a face from its thermal image using the method proposed by Frangi \textit{et al.}\ \cite{FranNiesVincVier1998}. Their so-called vesselness filter, first proposed for use on 3D MRI data, extracts tubular structures from an image. For a 2D image consider the two eigenvalues $\lambda_1$ and $\lambda_2$ of the Hessian matrix computed at a certain image locus and at a particular scale. Without loss of generality let us also assume that $|\lambda_1| \leq |\lambda_2|$. The two key values used to quantify how tubular the local structure at this scale is are $\mathcal{R}_\mathcal{A} = |\lambda_1|/|\lambda_2|$ and $\mathcal{S} = \sqrt{\lambda_1^2 + \lambda_1^2}$. The former of these measures the degree of local 'blobiness'. If the local appearance is blob-like, the Hessian is approximately isotropic and $|\lambda_1|\approx|\lambda_2|$ making $\mathcal{R}_\mathcal{A}$ close to 1. For a tubular structure $\mathcal{R}_\mathcal{A}$ should be small. On the other hand, $\mathcal{S}$ ensures that there is sufficient local information content at all: in nearly uniform regions, both eigenvalues of the corresponding Hessian will have small values. For a particular scale of image analysis $s$, the two measures, $\mathcal{R}_\mathcal{A}$ and $\mathcal{S}$, are then unified into a single vesselness measure:
{\begin{align}
  \mathcal{V}(s) =
    \begin{cases}
      0 &~ \text{if } \lambda_2 > 0\\
      (1-e^{-\frac{\mathcal{R}_\mathcal{B}}{2\beta^2}}) \times (1-e^{-\frac{\mathcal{S}}{2c^2}}) &~ \text{otherwise},
    \end{cases}
\end{align}}
where $\beta$ and $c$ are the parameters that control the sensitivity of the filter to $\mathcal{R}_\mathcal{A}$ and $\mathcal{S}$. Finally, if an image is analyzed across scales from $s_{min}$ to $\leq s_{max}$, the vesselness of a particular image locus can be computed as the maximal vesselness across the range:
{\begin{align}
  \mathcal{V}_0 = \max_{s_{min} \leq s \leq s_{max}} \mathcal{V}(s).
\end{align}}
Vesselness at three different scales for an example thermal image is illustrated in Fig.~\ref{f:frangi}(a-c), and the corresponding multi-scale result in Fig.~\ref{f:frangi}(d).

\subsubsection{Matching}\label{sss:matching}
\begin{figure}[thp]
  \centering
  \subfigure[From frontal reference ]
    {~~~~~~~~~~\includegraphics[height=0.2\textwidth,trim= 0mm 0mm 0mm 0mm,clip]{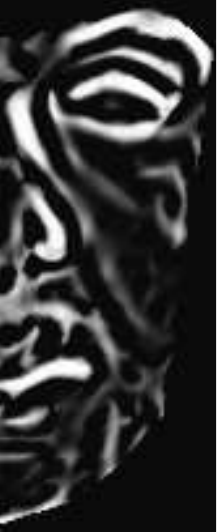}~~~~~~~~~~}
  \subfigure[From novel input at $70^\circ$ yaw]
    {~~~~~~~~~~\includegraphics[height=0.2\textwidth,trim=22mm 0mm 0mm 0mm,clip]{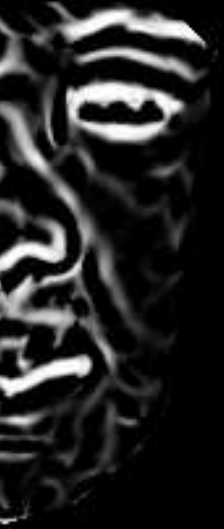}~~~~~~~~~~}
  \caption{ Examples of two normalized (warped) vascular networks extracted from thermal IR images of the same person's face in vastly different poses: (a) frontal ($0^\circ$ yaw), and (b) between semi-profile and full profile (yaw of approximately $70^\circ$). Notice the importance of the smoothness of our representation i.e.\ of our vesselness encoding by confidence rather than a binary decision. As illustrated on this example (and quantitatively demonstrated in Sec.~\ref{s:eval}) this ensures the preservation of the discriminative features of a face. In contrast, as witnessed by the change in the exact responses of the vesselness filter, had binarization been performed many important features would have been lost.}
  \label{f:vnexample}
\end{figure}

After the fitting of an AAM to the gallery images and the novel face, all of them, or indeed the extracted vesselness signatures, can be warped to a canonical frame. This warping normalizes data with respect to pose and facial expression, while the underlying representation ensures robustness with respect to various extrinsic factors which affect the face temperature (absolute as well as relative across different parts of the face, as discussed in `Multi-scale blood vessel extraction'). An example of two normalized vascular networks extracted from images of the same face in drastically different poses is shown in Fig.~\ref{f:vnexample}. Since the vesselness image inherently exhibits small shift invariance due to the smoothness of the vesselness filter response, as readily observed on examples in Fig.~\ref{f:vnexample} as well as in Fig.~\ref{f:frangi}(c), thus normalized images can be directly compared. In this paper we adopt the simple cross-correlation coefficient as a measure of similarity. If $I_{n1}$ and $I_{n2}$ are two normalized images (warped vesselness signatures), their similarity $\rho$ is computed as:
{\begin{align}
  \rho(I_{n1},&I_{n2}) = \label{e:dist} \\
  &\frac{ \sum_{i,j} (I_{n1}(i,j)- \bar{I}_{n1}) (I_{n2}(i,j)- \bar{I}_{n2}) } { \sqrt{\sum_{i,j} (I_{n1}(i,j)- \bar{I}_{n1})^2 \times \sum_{i,j} (I_{n2}(i,j)- \bar{I}_{n2})^2}}, \notag
\end{align}}
where $\bar{I}_{n1}$ and $\bar{I}_{n2}$ are the mean values of the corresponding images.

\section{Experimental evaluation}\label{s:eval}
In this section we report our empirical evaluation of the methods proposed in this paper. We start by describing the data set used in our experiments, follow with an explanation of the adopted evaluation protocol, and finish with a report of the results and their discussion.

\subsection{Data sets}\label{ss:data}

\subsubsection{University of Houston data set}\label{sss:uh}
We chose to use the University of Houston data set for our experiments. There are several reasons for this choice. Firstly, this is one of the largest data sets of thermal images of faces; it contains data from a greater number of individuals than Equinox \cite{DbEqui}, IRIS \cite{DbIRIS} or Florida State University \cite{SrivLiu2003} collections, and a greater variability in pose and expression than those of University of Notre Dame \cite{DbUND} or the University of California/Irvine \cite{PanHealPrasTrom2005}. Secondly, we wanted to make our results directly comparable to those of Buddharaju \textit{et al.}\ whose method bears the most resemblance to ours in spirit (but not in actual technical detail).

\begin{figure*}[htb]
  \centering
  \includegraphics[width=0.16\textwidth]{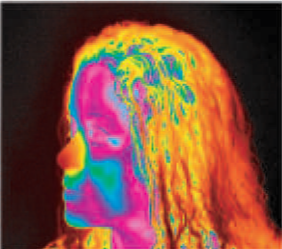}\hspace{10pt}
  \includegraphics[width=0.16\textwidth]{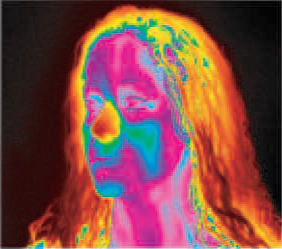}\hspace{10pt}
  \includegraphics[width=0.16\textwidth]{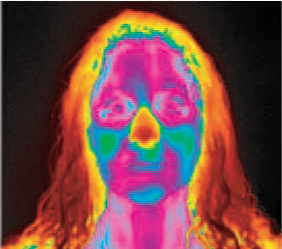}\hspace{10pt}
  \includegraphics[width=0.16\textwidth]{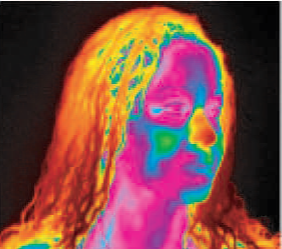}\hspace{10pt}
  \includegraphics[width=0.16\textwidth]{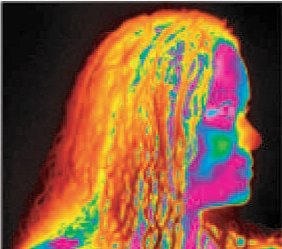}
  \caption{ False colour thermal appearance images of a subject in the five key poses in the University of Houston data set. }
  \label{f:dbUH}
  \vspace{13pt}
\end{figure*}

The University of Houston data set consists of a total of 7590 thermal images of 138 subjects, with a uniform number of 55 images per subject. The ethnicity, age and sex of subjects vary across the database. With the exception of four subjects, from whom data was collected in two sessions six months apart, the data for a particular subject was acquired in a single session. The exact protocol which was used to introduce pose and expression variability in the data set was not described by the authors \cite{BuddPavlTsiaBaza2007}. Example images are shown in Fig.~\ref{f:dbUH}. The database is available free of charge upon request.

\subsubsection{Thermal IR face motion data set}\label{sss:facemotionDB}
Although it is the largest and the most challenging IR face data set available publicly, as we report in Sec.~\ref{ss:res}, on this database the proposed method not only outperformed the previous state of the art but it also achieved perfect performance. In the interest of evaluating our method on a more challenging task, we collected a novel data set with an even greater and less constrained amount of pose and facial expression variation. In addition, whenever possible we attempted to capture the same subject after a time lapse and after exposure to different environmental conditions (outside temperature specifically). The present paper is the first and indeed the only work that has used this data.

\begin{figure*}[htb]
  \centering
  \includegraphics[width=0.91\textwidth]{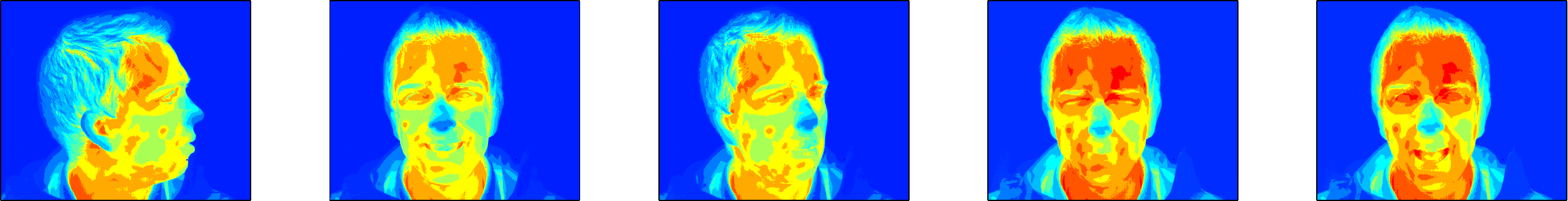}
  \caption{ False colour thermal appearance images of a subject in five arbitrary poses and facial expressions in the Thermal IR Face Motion data set
                  acquired by the authors. The collected Thermal IR Face Motion data set will be made freely publicly available following the publication of the present work. }
  \label{f:dbFM}
\end{figure*}

Each video sequence in the database is 10~s long and was captured at 30~fps, thus resulting in 300 frames of $320 \times 240$ pixels. The imaged subjects were instructed to perform head motion that covers the yaw range from frontal ($0^\circ$) to approximately full profile ($\pm 90^\circ$) face orientation relative to the camera, without any special attention to the tempo of the motion or the time spent in each pose. The subjects were also asked to display an arbitrary range of facial expressions. Examples of frames from a single video sequence in our data set is shown in Fig.~\ref{f:dbFM}.

The collected Thermal IR Face Motion data set will be made freely publicly available following the publication of our work.

\subsection{Evaluation methodology}\label{ss:evalMethod}
We evaluated the proposed algorithm in a setting in which the algorithm is trained using only a single image in an arbitrary pose and facial expression. The querying of the algorithm using a novel face is also performed using a single image, in a different pose and/or facial expression. Both pose and facial expression changes present a major challenge to the current state of the art, and the consideration of the two in combination make our evaluation protocol extremely challenging (indeed, more so than any attempted by previous work), and, importantly, representative of the conditions which are of interest in a wide variety of practical applications.

\subsection{Results and discussion}\label{ss:res}
\paragraph{University of Houston data set}
To assess the performance of the proposed method, we first examined its rank-$N$ (for $N=1, \ldots$) recognition rate, i.e.\ its cumulative match characteristics, on the University of Houston data set. Our method was found to exhibit perfect performance at rank-1 already, correctly recognizing all of the subjects in the database. This result is far superior to the previously proposed thermal minutia points based approach of Buddharaju \textit{et al.} \cite{BuddPavlTsia2006} which correctly recognizes 82.5\% of the individuals at rank-1 and does not reach 100\% even for rank-20 matching, or indeed the iterative vascular network registration based method of Pavlidis and Buddharaju \cite{PavlBudd2009} which correctly recognizes 96.2\% of the individuals rank-1 and which also fails to achieve 100\% even at rank-20.

The performance of the proposed method is all the more impressive when it is considered that unlike the aforementioned previous work, we perform recognition using a single training image only, and across pose and facial expression changes. Both Buddharaju \textit{et al.}, and Pavlidis and Buddharaju train their algorithm using multiple images. In addition, it should be emphasized that they do not consider varying pose -- the pose of an input face is first categorized by pose and then matched with training faces in that pose only. In contrast, we perform training using a single image only, truly recognizing \emph{across} pose variations. A comparative summary is shown in Tab.~\ref{t:res}.

\begin{table*}
  \centering
  \renewcommand{\arraystretch}{1.2}
  \vspace{5pt}
  \small
  \caption{ A summary of the key evaluation results and method features of the proposed algorithm, and the two previously proposed vascular network based approaches of Buddharaju \textit{et al.}~\cite{BuddPavlTsia2006}, and Pavlidis and Buddharaju~\cite{PavlBudd2009}. Legend: $\CIRCLE$ large degree of invariance; $\RIGHTcircle$ some degree of invariance; $\Circle$ little to no invariance.  }

  \begin{tabular}{l||ccc|ccccc}
    \Hline
                                 & \multicolumn{3}{c}{Recognition rate} & \multicolumn{4}{c}{Invariance to}\\
    \cline{2-9}
                                 & \multirow{2}{*}{Rank-1} & \multirow{2}{*}{Rank-3}  & \multirow{2}{*}{Rank-5}  & \multirow{2}{*}{Expression}     & \multicolumn{2}{c}{Pose changes}           & Physiological & \multirow{2}{*}{Scale} \\
                                 &        &         &         &                & (small) & (large)              & condition     &       \\
    \hline
    AAM + multi-scale vesselness & \multirow{2}{*}{100.0\%} & \multirow{2}{*}{100.0\%} & \multirow{2}{*}{100.0\%} & \multirow{2}{*}{$\CIRCLE$}      & \multirow{2}{*}{$\CIRCLE$}      & \multirow{2}{*}{$\CIRCLE$} & \multirow{2}{*}{$\CIRCLE$}   & \multirow{2}{*}{$\CIRCLE$}\\
    (the proposed method)&        &         &         &                &                &      &       \\
    vascular network alignment    &  \multirow{2}{*}{96.2\%} & \multirow{2}{*}{98.3\%}  &  \multirow{2}{*}{99.0\%} & \multirow{2}{*}{$\RIGHTcircle$} & \multirow{2}{*}{$\RIGHTcircle$} & \multirow{2}{*}{$\Circle$} & \multirow{2}{*}{$\CIRCLE$}    & \multirow{2}{*}{$\Circle$}\\
      \cite{PavlBudd2009}&        &         &         &                &                &      &       \\
    thermal minutiae points  &  \multirow{2}{*}{82.5\%} & \multirow{2}{*}{92.2\%}  &  \multirow{2}{*}{94.4\%} & \multirow{2}{*}{$\RIGHTcircle$} & \multirow{2}{*}{$\RIGHTcircle$} & \multirow{2}{*}{$\Circle$} & \multirow{2}{*}{$\CIRCLE$}    & \multirow{2}{*}{$\Circle$}\\
    \cite{BuddPavlTsia2006}&        &         &         &                &                &      &       \\
    \Hline
  \end{tabular}
  \label{t:res}
  \vspace{12pt}
\end{table*}

\paragraph{IR face motion data set}
Following the evaluation of our method on the University of Houston data set, we investigated its performance on our newly acquired thermal IR face motion database. The more precise control over the acquisition process and the nature of variability present in this data allowed us to conduct a more systematic and in-depth analysis, while at the same time the greater extent of variability within the data ensured a greater level of challenge of the experiment in comparison to the previous one.

\begin{figure*}[htb]
  \centering
  \subfigure[ROC, average]{\includegraphics[height=0.22\textwidth]{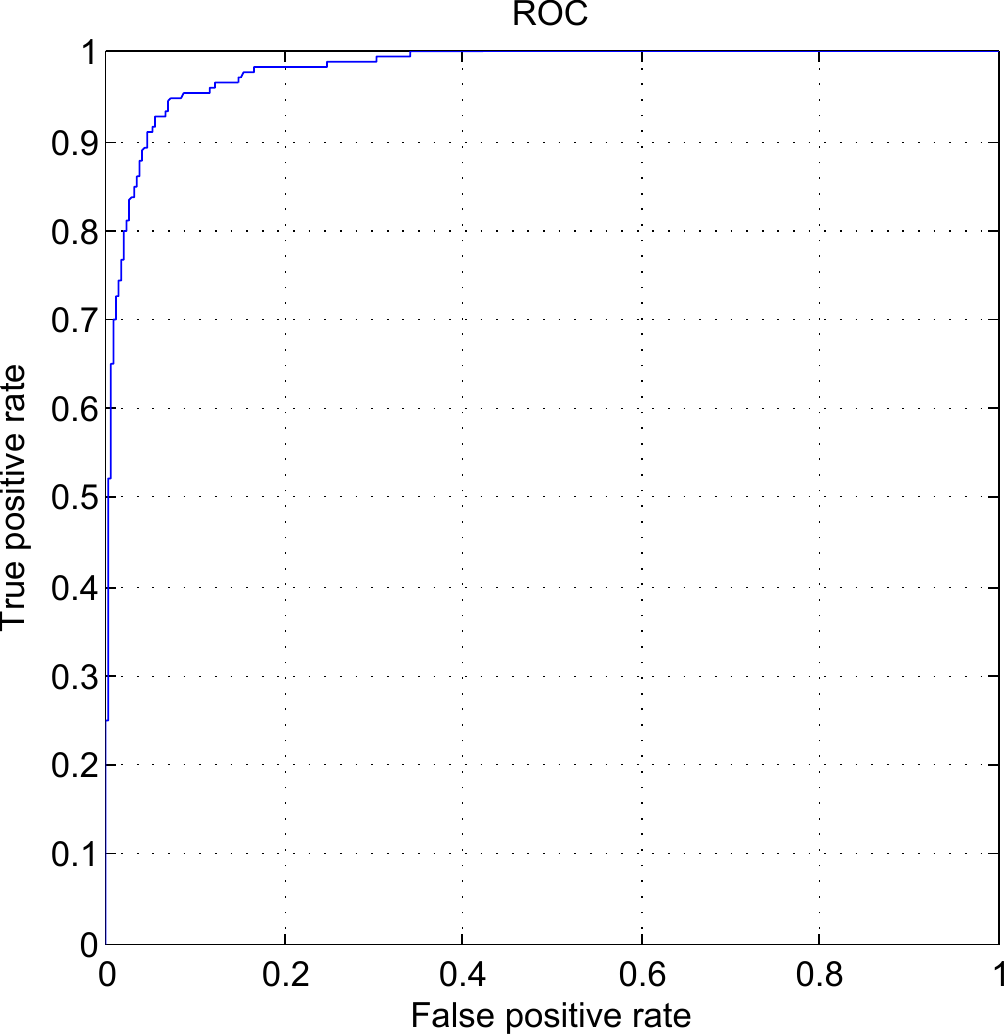}}~~~~
  \subfigure[ROC, $\leq 30^\circ$ yaw change at yaw $0-30^\circ$]{\includegraphics[height=0.22\textwidth]{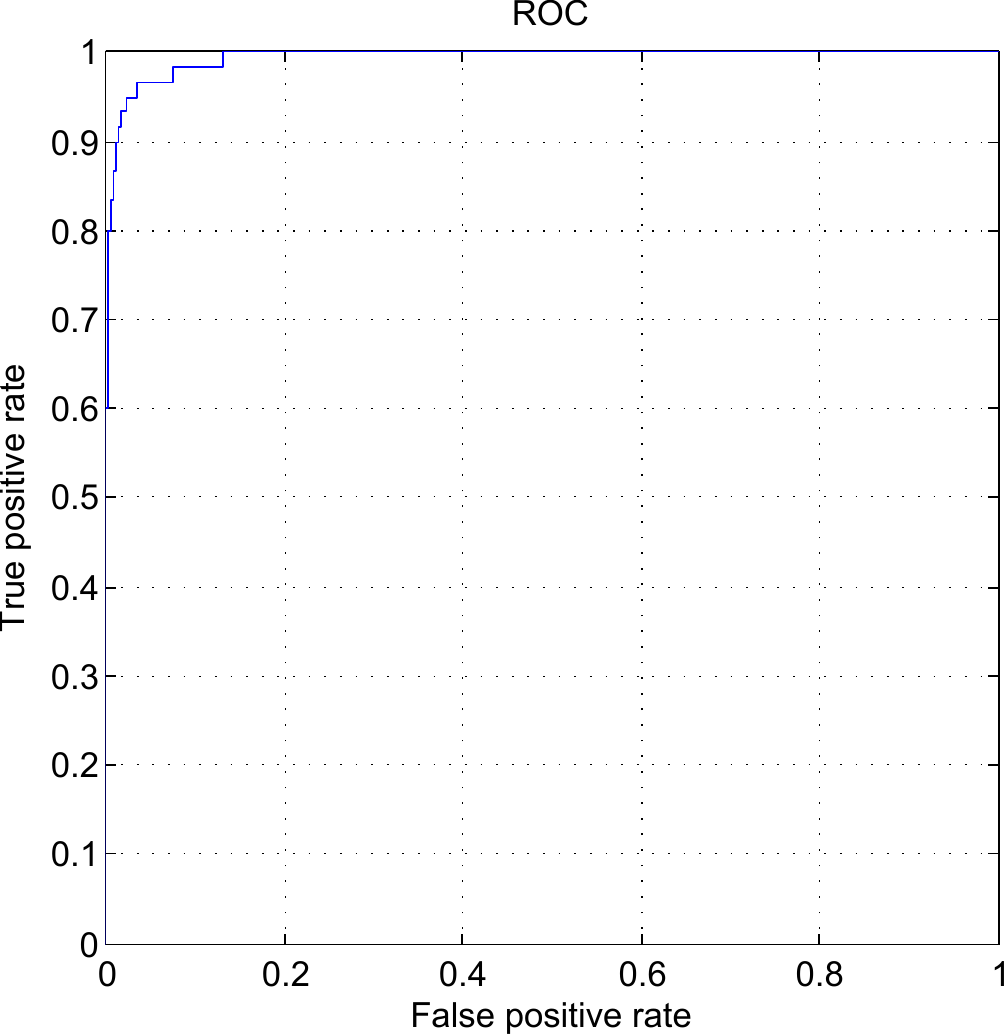}}~~~~
  \subfigure[ROC, $\leq 30^\circ$ yaw change at yaw $30-60^\circ$]{\includegraphics[height=0.22\textwidth]{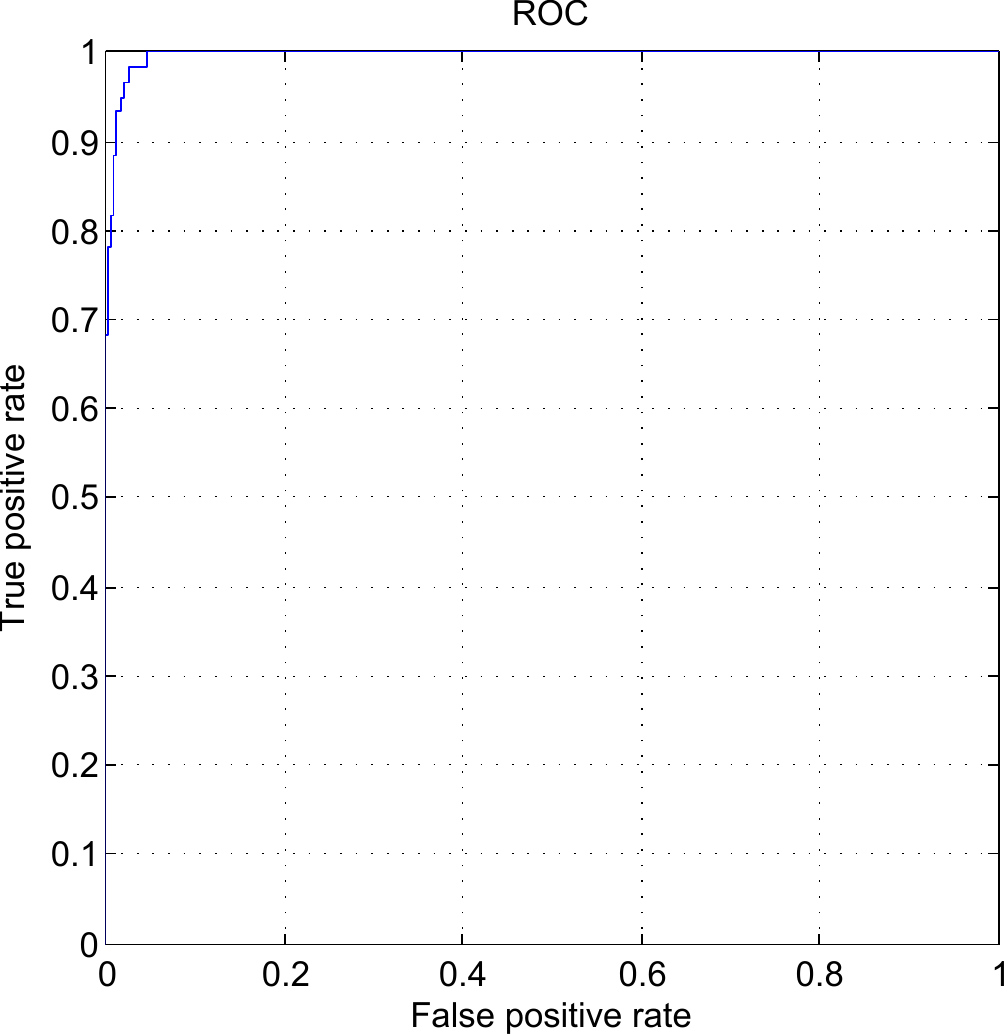}}~~~~
  \subfigure[ROC, $\leq 30^\circ$ yaw change at yaw $60-90^\circ$]{\includegraphics[height=0.22\textwidth]{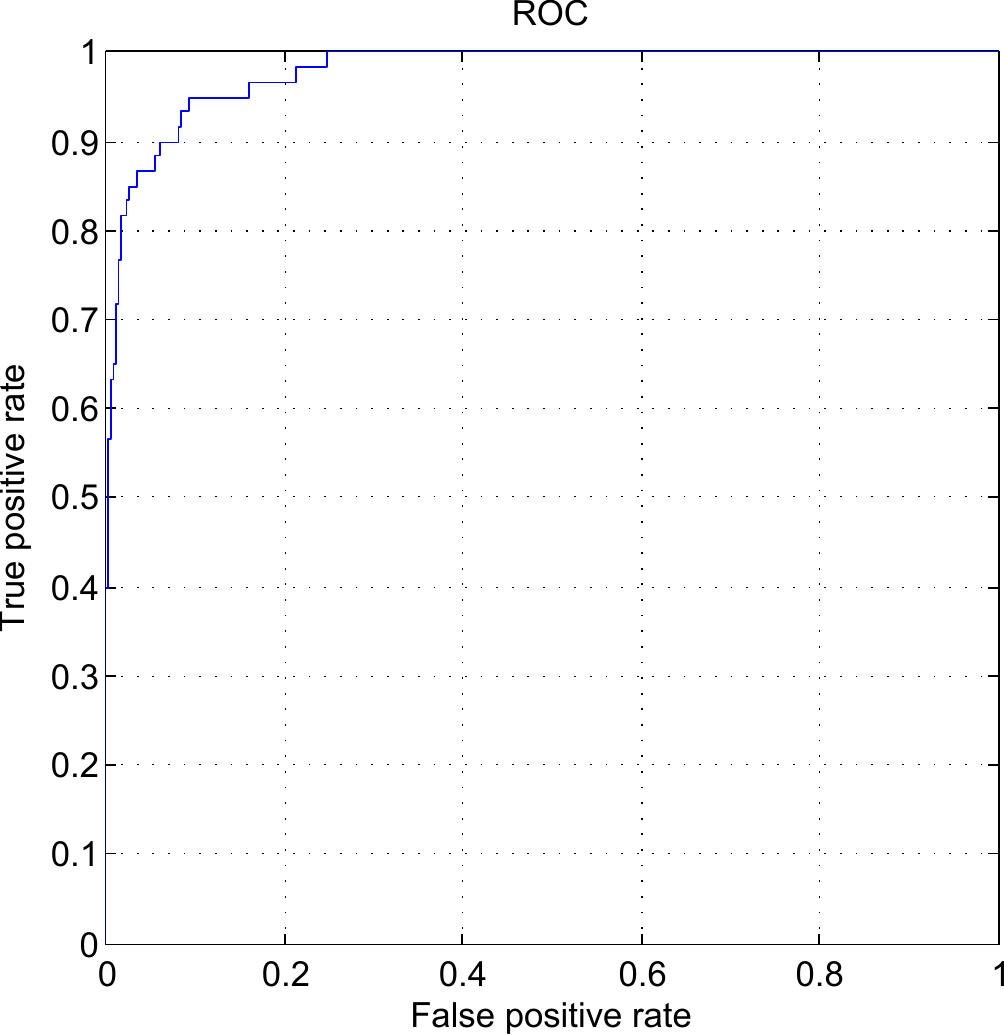}}
  \caption{ The receiver-operator characteristic (ROC) curves achieved by the proposed method on the Thermal IR Face Motion Database (see Sec.~\ref{sss:facemotionDB} and Fig.~\ref{f:dbFM}). Shown are (a) the average ROC curve over the entire differential yaw range of $0-90^\circ$ (all differential yaw values being equally represented in the evaluation), and the three ROC curves for yaw changes of up to $\pm 15^\circ$ at three intervals of yaw: (b) $0-30^\circ$, (c) $30-60^\circ$, and (d) $60-90^\circ$. Notice that the head orientations around the semi-profile pose proved to be more discriminative than those near the fully frontal face, as witnessed the ROC curve in (c) being closer to the ideal characteristic than those in (b) or (d).  }
  \label{f:roc}
\end{figure*}

\begin{figure}[htb]
  \centering
  \includegraphics[width=0.48\textwidth]{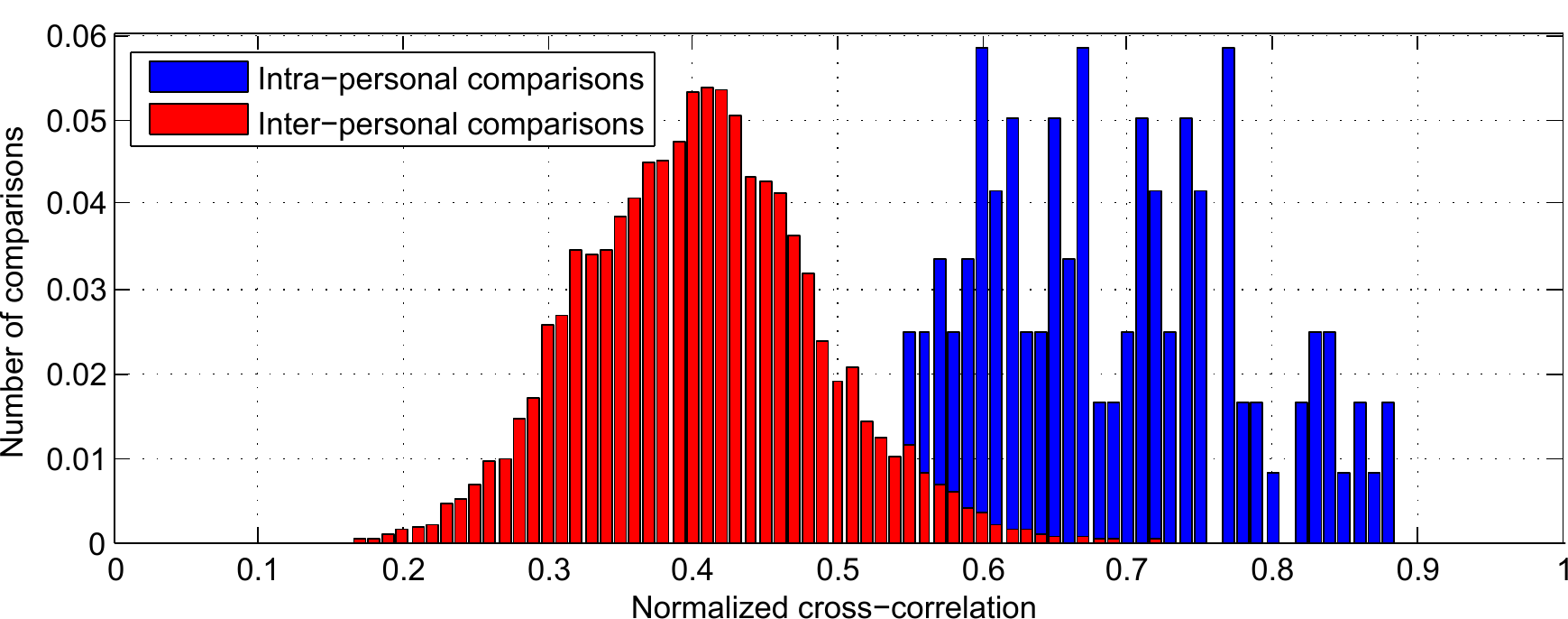}
  \caption{ Separation of inter-class (red) and intra-class (blue) similarities (see Eq~\eqref{e:dist} in Sec.~\ref{sss:matching}) on the Thermal IR Face Motion data set, shown as plots of the corresponding histograms (normalized to unit sum, for easier visualization). }
  \label{f:hists}
\end{figure}

Our algorithm again achieved the perfect recognition rate already at rank-1. The corresponding receiver-operator characteristic (ROC) curve is shown in Fig.~\ref{f:roc}(a). The separation of classes is further illustrated by the plot in Fig.~\ref{f:hists} which shows the observed separation inter-class and intra-class distances (see Eq~\eqref{e:dist} in Sec.~\ref{sss:matching}). Both plots in Fig.~\ref{f:roc}(a) and Fig.~\ref{f:hists} correspond to the average performance of the proposed method across pose changes, all yaw differentials being equally represented in the unseen data used to query the algorithm. To gain further insight, we also examined the performance of our approach at different yaw ranges. Specifically, we measured the recognition performance constrained to yaw changes of up to $\pm 15^\circ$ at three intervals of yaw: $0-30^\circ$, $30-60^\circ$, and $60-90^\circ$. Our results are consistent with previous reports in the literature on visible spectrum face recognition. The most difficult poses were those close to the full profile. Likely explanations for this include inherent reasons, such as the absence of information redundancy and out of plane geometric information (via self-occlusion), as well as those of a practical nature, such as the increased difficulty of accurate AAM fitting. Interestingly, much like in the visible spectrum \cite{SimZhang2004,AranCipo2013}, head orientations around the semi-profile pose proved to be more discriminative than those near the fully frontal face. This finding may have interesting implications to the design and deployment of thermal IR based face recognition systems in practice.

\section{Summary and Conclusions}\label{s:conc}
In this paper we described a novel method for face recognition using thermal IR images. Our work addressed two main challenges. These are the variations of thermal IR appearance due to (i) change in head pose and expression, and (ii) facial heat pattern emissions (e.g.\ effected by ambient temperature or sympathetic nervous system activity).

When comparing two thermal IR images of faces, we normalize their poses and facial expressions by using an active appearance model to generate synthetic images of the two faces in the same view (corresponding to the average of the two input poses). A major contribution of our work is the use of an AAM ensemble in which each AAM is specialized to a particular range of poses \emph{and} a particular region of the thermal IR face space. In addition, we show how AAM convergence problems associated with the lack of high frequency detail in thermal images can be overcome by a pre-processing stage which enhances discriminative information content.

To achieve robustness to changes in facial heat pattern emissions, we describe a representation which is not based on either absolute or relative facial temperature but instead unchangeable anatomic features in the form of a subcutaneous vascular network. We introduce a more robust vascular network extraction than that used in the literature to date. Our approach is based on the so-called vesselness filter. This method allows us to process the face in multi-scale fashion and account for the confidence that a particular image locus corresponds to a vessel, thus achieving greater resilience to head pose changes, face scale and input image resolution. The effectiveness of the proposed algorithm was demonstrated on the largest publicly available data set, which includes large pose and facial expression variation, and a novel data set acquired by the authors (which will be made freely available), which comprises thermal IR video sequences of people performing head motion and changing facial expressions. Perfect performance was achieved on both data sets.

\paragraph{Future work}
The proposed framework readily opens avenues for further research and improvement. Our first aim in the short term is to investigate the use of non-parametric methods (e.g.\ hierarchial Dirichlet process based methods \cite{TehJordBealBlei2004}) for clustering data used to train individual AAMs in our ensemble (see Sec.~\ref{ss:ensemble}). This, we suspect, may be of particular benefit when our method is applied to databases with large numbers of individuals. In the longer term, our goal is to extend the proposed method to be able to cope automatically with constrained (in spatial as well as appearance sense) occlusions, such as facial hair or eyeglasses (which are opaque to the wavelengths in the LWIR spectrum). The use of partial AAMs appears promising in this regard, and we expect that the techniques we introduced in order to increase the accuracy and reliability of the AAM fitting process (see Sec.~\ref{ss:ensemble} and~\ref{ss:preprocessing}) will prove to be of much value in this context.

\footnotesize
\bibliographystyle{ieeetran}
\bibliography{./my_bibliography,./eccv02}

\end{document}